\documentclass[conference]{IEEEtran}

\usepackage{cite}
\usepackage{amsmath,amssymb,amsfonts}
\usepackage{graphicx}
\usepackage{textcomp}
\usepackage{xcolor}
\usepackage{hyperref}
\usepackage{cleveref}
\usepackage{xcolor} 
\usepackage{tikz}
\usetikzlibrary{arrows.meta,positioning}
\usepackage{booktabs}
\usepackage{tabularx}
\usepackage{makecell}
\usepackage{listings}
\usepackage{adjustbox}
\usepackage{subcaption}
\definecolor{codegreen}{rgb}{0,0.6,0}
\definecolor{codegray}{rgb}{0.5,0.5,0.5}
\definecolor{codepurple}{rgb}{0.58,0,0.82}
\definecolor{backcolour}{rgb}{0.96,0.96,0.96}
\usepackage{microtype}

\def\BibTeX{{\rm B\kern-.05em{\sc i\kern-.025em b}\kern-.08em
    T\kern-.1667em\lower.7ex\hbox{E}\kern-.125emX}}

\usepackage{microtype}

\title{Compiling VGDL into Causal Models}

\author{
  \IEEEauthorblockN{Mohit Jiwatode} 
  \IEEEauthorblockA{\textit{Institute for Information Processing}\\
  \textit{Leibniz University Hannover}\\
  Hannover, Germany \\
  jiwatode@tnt.uni-hannover.de}
  \and
  \IEEEauthorblockN{Bodo Rosenhahn} 
  \IEEEauthorblockA{\textit{Institute for Information Processing}\\
  \textit{Leibniz University Hannover}\\
  Hannover, Germany \\
  rosenhahn@tnt.uni-hannover.de}
  \and
  \IEEEauthorblockN{Alexander Dockhorn} 
  \IEEEauthorblockA{\textit{SDU Metaverse Lab}\\
  \textit{University of Southern Denmark}\\
  Odense, Denmark \\
  adoc@sdu.dk}
}

\begin{document}

\maketitle

\begin{abstract}


Reinforcement learning and large language models often struggle to accurately capture the causal mechanics of game environments. Standard reinforcement learning agents tend to rely on spurious correlations, while large language models are prone to hallucinating game rules. Although causal reinforcement learning improves interpretability, there is currently no formal methodology to map complex game mechanics directly into causal models.
To address this, we propose a deterministic framework that compiles games specified in the Video Game Description Language into Dynamic Structural Causal Models. Rather than inferring causal structures from gameplay traces or noisy large language models' outputs, our methodology directly translates game components, including sprite dynamics, interaction rules, and termination conditions, into explicit structural equations. Each game tick represents a causal transition from state variables at time $t$ to $t+1$.
By establishing this grounded mapping, the approach guarantees absolute causal fidelity to the ground-truth game mechanics. The resulting models offer transparent causal pathways that support counterfactual reasoning, causal reinforcement learning agent training, and procedural content validation. This framework provides a principled bridge between symbolic game descriptions and causally grounded game AI.

\end{abstract}

\begin{IEEEkeywords}
GVGAI, VGDL, Large Language Models, Causality, Causal Reinforcement Learning
\end{IEEEkeywords}

\section{Introduction}

Reinforcement learning (RL) is widely used in modern game AI, but standard methods struggle to understand underlying causal game mechanics. The deep learning models used in RL are generally opaque, which limits the interpretability of their decisions~\cite{deng2023causal}. Games frequently involve hidden rules or changing environments (e.g., new enemies or game phases) which pose adaptation challenges for naive learners~\cite{li2026confounding}. To make such rules explicit to an agent, Apeldoorn et al.~\cite{Apeldoorn2021Exception} have used hierarchical knowledge bases that are updated while playing the game. As those are learned by observation, they may include spurious correlations and therefore not represent the true rules of the underlying environment. Causal RL addresses these core problems by filtering out spurious correlations and focusing on causally relevant information, while also remaining interpretable~\cite{deng2023causal}. Recent studies~\cite{madumal2020explainable} further prove that causal explanations of RL agents significantly improve user understanding and trust. 

Despite this promise, existing game benchmarks and agents rarely incorporate causality explicitly, and there exists no formal framework for mapping game mechanics into causal models. While logic-based game description languages (e.g., GDL and systems like Ludocore~\cite{smith2010ludocore}) have supported strategy analysis and logical forms of causal reasoning ~\cite{genesereth2022analyzing}, they do not provide the type of structured causal models targeted here. 
Similarly, large language models (LLMs) applied to game reasoning suffer from hallucinations, misunderstand spatial relationships, and often perform subpar in gameplay~\cite{li2025gvgai,jiwatode_spatial_reasoning}. Although incorporating causal models can improve their understanding of game mechanics~\cite{jiwatode2026gameplay}, LLMs may still misidentify such models, and without ground truth, verification is often intractable. We address this gap by introducing a formal mapping framework that equips AI agents with deterministic causal models, enabling more efficient learning through grounded causal information.


\section{Background and Related Work}

\subsection{Causality}

Causality studies cause–and–effect relationships, enabling models to predict outcomes rather than just correlations~\cite{pearl2009causality}. Pearl’s ladder of causation defines three levels: association (statistical correlations), intervention (actively setting variables independent of their causes), and counterfactuals (reasoning about alternative outcomes to observed events)~\cite{pearl2009causality}.

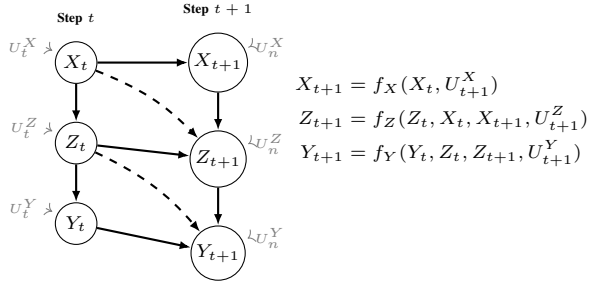
\begin{figure}[t]
\centering
\begin{tikzpicture}[
    node distance=5mm and 12mm,
    var/.style={circle, draw, minimum size=5.5mm, font=\scriptsize, inner sep=1pt},
    arr/.style={-{Latex[length=1.5mm]}, thick},
    lag/.style={-{Latex[length=1.5mm]}, thick, dashed},
    noise/.style={shorten <=2pt, <-, font=\tiny, color=gray}
]

\node[var] (Xt) {$X_t$};
\node[var, below=of Xt] (Zt) {$Z_t$};
\node[var, below=of Zt] (Yt) {$Y_t$};

\node[var, right=of Xt] (Xn) {$X_{t+1}$};
\node[var, below=of Xn] (Zn) {$Z_{t+1}$};
\node[var, below=of Zn] (Yn) {$Y_{t+1}$};

\node[above=1mm of Xt, font=\tiny\bfseries] {Step $t$};
\node[above=1mm of Xn, font=\tiny\bfseries] {Step $t+1$};

\draw[arr] (Xt) -- (Zt);
\draw[arr] (Zt) -- (Yt);
\draw[arr] (Xn) -- (Zn);
\draw[arr] (Zn) -- (Yn);
\draw[arr] (Xt) -- (Xn);
\draw[arr] (Zt) -- (Zn);
\draw[arr] (Yt) -- (Yn);
\draw[lag] (Xt) to[bend left=12] (Zn);
\draw[lag] (Zt) to[bend left=12] (Yn);

\draw[noise] (Xt) -- ++(-0.4,0.2) node[left=-1pt] {$U^X_t$};
\draw[noise] (Zt) -- ++(-0.4,0.2) node[left=-1pt] {$U^Z_t$};
\draw[noise] (Yt) -- ++(-0.4,0.2) node[left=-1pt] {$U^Y_t$};
\draw[noise] (Xn) -- ++(0.4,0.2) node[right=-1pt] {$U^X_{n}$};
\draw[noise] (Zn) -- ++(0.4,0.2) node[right=-1pt] {$U^Z_{n}$};
\draw[noise] (Yn) -- ++(0.4,0.2) node[right=-1pt] {$U^Y_{n}$};

\node[right=5mm of Xn, anchor=north west, font=\scriptsize] (eqs) {
$\begin{aligned}
X_{t+1} &= f_X(X_t,U^X_{t+1}) \\
Z_{t+1} &= f_Z(Z_t,X_t,X_{t+1},U^Z_{t+1}) \\
Y_{t+1} &= f_Y(Y_t,Z_t,Z_{t+1},U^Y_{t+1})
\end{aligned}$
};

\end{tikzpicture}
\caption{Dynamic SCM: Graph (left) showing causal dependencies and structural equations (right) defining recursive assignments.}
\label{fig:dynamic_scm}
\end{figure}

A Structural Causal Model (SCM) is a triple $(U,V,F)$ where $U$ is a set of exogenous variables, $V$ a set of endogenous variables, and $F=\{f_v\}_{v\in V}$ a set of functions deriving each $v\in V$ from its parents $\mathrm{Pa}(v)\subseteq V$ and corresponding noise variable $U_v$. SCMs allow for a systematic derivation of intervention effects and therefore facilitate the design of algorithms that exploit causal structures in complex domains~\cite{pearl2009causality}. Furthermore, a Dynamic Structural Causal Model (DSCM) extends this framework to temporal sequences by indexing variables over discrete time steps $t \in \mathcal{T}$~\cite{pearl2009causality}. In a DSCM, the functions $F$ determine the state of $v_t \in V_t$ based on its parents in current and preceding time steps, $\mathrm{Pa}(v_t) \subseteq V_{\le t}$, thereby capturing the causal dependencies over time~\cite{pearl2009causality}. An example of a DSCM is shown in ~\Cref{fig:dynamic_scm}.
Madumal et al.~\cite{madumal2020explainable} use SCMs to explain RL agent behavior, generating causal explanations by analyzing counterfactuals.  Hammond et al.~\cite{hammond2023reasoning} extend causal modeling to games, proposing ``causal games'' that encode agents' strategies and dependencies in a unified framework.  These works illustrate how causal models offer insight into decision-making
and prediction 
beyond correlations.

\subsection{General Video Game AI (GVGAI) and Video Game Description Language (VGDL)}
The GVGAI framework~\cite{gvgaibook2019} allows comparing the performance of AI agents in playing a large corpus of arcade-like video games using a unified interface. 
Each game is defined in VGDL, a lightweight, human-readable language format consisting of a sprite set, a level mapping, a termination set, and an interaction set. While the sprite set describes all components of a game, the termination and interaction set describes the game's logic. The latter does so by describing the outcome of collision events in terms of cause and effect relationships, effectively representing a high-level SCM.

Our approach models a video game as a \emph{dynamic} SCM: a two-time-slice graph in which variables at time $t$ influence those at $t+1$.  Prior work by Jiwatode et al.~\cite{jiwatode2026gameplay} and Chen et al.~\cite{chen2025causal} integrates SCMs with LLMs for causal induction in diverse environments.  We take a complementary route: rather than prompt an LLM to predict rules, we deterministically compile the known VGDL rules into an SCM.  This guarantees causal fidelity (the model matches the ground-truth game mechanics) without relying on noisy language-model inference.

\section{Dynamic Causal Model of Games}


We model a VGDL game as a Dynamic SCM (DSCM), focusing on mapping VGDL directly to SCM components rather than learning correlations from traces~\cite{jiwatode2026gameplay}. While VGDL encodes these elements compactly, formalizing them as an SCM enables causal RL, interventions, and counterfactual analysis, which VGDL alone does not support for decision-making. Each game tick is treated as a causal transition from state $t$ to $t+1$, where sprite dynamics, layout, interactions, and termination rules define the causal structure.

\begin{figure}[t]
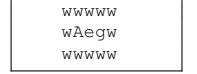

\centering

\begin{subfigure}[t]{0.65\linewidth} 
\centering
\begin{adjustbox}{max width=\linewidth}
\begin{minipage}{\linewidth}
\begin{lstlisting}[
    basicstyle=\ttfamily\scriptsize,
    breaklines=true,
    columns=fullflexible,
    frame=single,
    xleftmargin=0.5em,
    xrightmargin=0.5em
]
BasicGame
    SpriteSet
        avatar > MovingAvatar
        wall   > Immovable
        goal   > Immovable
        enemy  > RandomNPC
    InteractionSet
        avatar wall  > stepBack
        goal avatar  > killSprite
        avatar enemy > killSprite
    LevelMapping
        A > avatar
        w > wall
        g > goal
        e > enemy
    TerminationSet
        SpriteCounter stype=goal limit=0 win=True
        SpriteCounter stype=avatar limit=0 win=False
\end{lstlisting}
\end{minipage}
\end{adjustbox}
\caption{VGDL specification}
\end{subfigure}
\hspace{0.5em} 
\begin{subfigure}[t]{0.12\textwidth} 
\centering
\vspace{0pt}
\begin{adjustbox}{max width=\linewidth}
\begin{minipage}{\linewidth}
\begin{lstlisting}[
    basicstyle=\ttfamily\scriptsize,
    frame=single,
    columns=fixed,
    framesep=2pt,     % Reduces internal padding between text and frame
    xleftmargin=1pt,  % Removes left white space
    xrightmargin=1pt, % Removes right white space
    aboveskip=0pt,    % Removes space above the listing
    belowskip=0pt     % Removes space below the listing
]
    wwwww
    wAegw
    wwwww
\end{lstlisting}
\end{minipage}
\end{adjustbox}
\caption{Level layout}
\end{subfigure}

\caption{\texttt{Basicgame} definition and example level layout}
\label{fig:basicgame-complete}
\end{figure}

Sprites are mapped to state representations defined by the dynamics of the selected games. 
We propose the following labels based on the possible descriptions of sprites in the \texttt{SpriteSet} in a VGDL: 
\texttt{Position}, \texttt{Orientation}, \texttt{Velocity}, \texttt{IsAlive}, \texttt{Resource}, \texttt{Health}, and \texttt{Time}. 
Not every selected game uses every attribute. 
For example, \texttt{basicgame} (see \Cref{fig:basicgame-complete}) uses position, orientation, velocity, alive-state, live-instance counts, and terminal-state variables, while resource and health variables are needed only for games whose VGDL dynamics include collection, inventory, damage, or health-based interactions. For a game tick $t$, the general endogenous state can be written as
\begin{equation}
\begin{aligned}
\mathcal{V}^{t} =
\{&P_s^t, O_s^t, \textit{Vel}_s^t, A_s^t, R_s^t, H_s^t, T^t,\\
  &Count_s^t, Terminal_k^t
  \;|\; s \in \mathcal{S},\ k \in \mathcal{K}\}.
\end{aligned}
\end{equation}

where $\mathcal{S}$ is the set of sprite types and instances, and $\mathcal{K}$ is the set of termination conditions. 
Here, $P_s^t$ denotes the position of sprite $s$, $O_s^t$ its orientation, $\textit{Vel}_s^t$ its velocity, $A_s^t$ whether the sprite is alive or active, $R_s^t$ its resource state, $H_s^t$ its health state, and $T^t$ the current game time or tick. 
The variable $Count_s^t$ denotes the number of live instances of a sprite type, and $Terminal_k^t$ denotes whether a win or loss condition has been reached. 
In \Cref{tab:basicgame-scm}, these variables are abbreviated as $P$, $O$, $V$, $A$, $Count$, and $Terminal$ to keep the mapping readable.

The exogenous variables include the player action, random choices, and the initial layout:
\begin{equation}
    \mathcal{U}^{t} =
    \{U_{\mathrm{action}}^t, U_{\mathrm{rng}}^t, Init_c^0\}.
\end{equation}
The variable $U_{\mathrm{action}}^t$ represents the action selected by the player or agent. 
The variable $U_{\mathrm{rng}}^t$ represents randomness used by stochastic sprite classes such as \texttt{RandomNPC}. 
The variable $Init_c^0$ represents the level-layout symbol at initialization. 
The layout is exogenous because it is given before play begins and determines which sprite instances exist and where they are placed. 
The VGDL and level layout for \texttt{basicgame} are shown in \Cref{fig:basicgame-complete}, and their corresponding SCM mapping is given in \Cref{tab:basicgame-scm}.

\begin{table*}[t]
\centering
\scriptsize
\caption{Concrete mapping for \texttt{basicgame} (see~\Cref{fig:basicgame-complete}).
Notation: $P$ = position, $O$ = orientation, $V$ = velocity, $A$ = alive, $Count$ = live-instance count, $Terminal$ = win/loss node, $U$ = exogenous input, RNG = random input.}
\label{tab:basicgame-scm}
\renewcommand{\arraystretch}{1.18}
\setlength{\tabcolsep}{5pt}
\begin{tabularx}{\textwidth}{@{}>{\raggedright\arraybackslash}p{2.5cm}>{\raggedright\arraybackslash}p{2.4cm}>{\raggedright\arraybackslash}p{5.7cm}>{\raggedright\arraybackslash}X@{}}
\toprule
\textbf{VGDL element} & \textbf{SCM nodes} & \textbf{SCM edges} & \textbf{Structural equation} \\[1pt]
\midrule

\multicolumn{4}{l}{\textbf{SpriteSet}} \\
\midrule
\texttt{avatar > MovingAvatar}
&
\makecell[tl]{$U_{\mathrm{action}}^t$, $P_{\mathrm{avatar}}^t$ \\[1pt]
$O_{\mathrm{avatar}}^t$, $V_{\mathrm{avatar}}^t$ \\[1pt]
$A_{\mathrm{avatar}}^t$, $P_{\mathrm{avatar}}^{t+1}$ \\[1pt]
$O_{\mathrm{avatar}}^{t+1}$, $V_{\mathrm{avatar}}^{t+1}$ \\[1pt]
$A_{\mathrm{avatar}}^{t+1}$}
&
\makecell[tl]{$U_{\mathrm{action}}^t \rightarrow P_{\mathrm{avatar}}^{t+1}$, $U_{\mathrm{action}}^t \rightarrow O_{\mathrm{avatar}}^{t+1}$ \\[1pt]
$U_{\mathrm{action}}^t \rightarrow V_{\mathrm{avatar}}^{t+1}$, $P_{\mathrm{avatar}}^t \rightarrow P_{\mathrm{avatar}}^{t+1}$ \\[1pt]
$O_{\mathrm{avatar}}^t \rightarrow P_{\mathrm{avatar}}^{t+1}$, $V_{\mathrm{avatar}}^t \rightarrow P_{\mathrm{avatar}}^{t+1}$ \\[1pt]
$A_{\mathrm{avatar}}^t \rightarrow A_{\mathrm{avatar}}^{t+1}$}
&
\makecell[tl]{$P_{\mathrm{avatar}}^{t+1} := f_{\mathrm{pos}}(P_{\mathrm{avatar}}^t, O_{\mathrm{avatar}}^t, V_{\mathrm{avatar}}^t, U_{\mathrm{action}}^t)$ \\[1pt]
$O_{\mathrm{avatar}}^{t+1} := f_{\mathrm{ori}}(O_{\mathrm{avatar}}^t, U_{\mathrm{action}}^t)$ \\[1pt]
$V_{\mathrm{avatar}}^{t+1} := f_{\mathrm{vel}}(V_{\mathrm{avatar}}^t, U_{\mathrm{action}}^t)$}
\\[1pt]
\specialrule{0.03em}{2pt}{2pt}

\texttt{wall > Immovable}
&
\makecell[tl]{$P_{\mathrm{wall}}^t$, $A_{\mathrm{wall}}^t$ \\[1pt]
$P_{\mathrm{wall}}^{t+1}$, $A_{\mathrm{wall}}^{t+1}$}
&
\makecell[tl]{$P_{\mathrm{wall}}^t \rightarrow P_{\mathrm{wall}}^{t+1}$, $A_{\mathrm{wall}}^t \rightarrow A_{\mathrm{wall}}^{t+1}$}
&
\makecell[tl]{$P_{\mathrm{wall}}^{t+1} := P_{\mathrm{wall}}^t$ \\[1pt]
$A_{\mathrm{wall}}^{t+1} := A_{\mathrm{wall}}^t$}
\\[1pt]
\specialrule{0.03em}{2pt}{2pt}

\texttt{goal > Immovable}
&
\makecell[tl]{$P_{\mathrm{goal}}^t$, $A_{\mathrm{goal}}^t$ \\[1pt]
$P_{\mathrm{goal}}^{t+1}$, $A_{\mathrm{goal}}^{t+1}$ \\[1pt]
$Count_{\mathrm{goal}}^{t+1}$}
&
\makecell[tl]{$P_{\mathrm{goal}}^t \rightarrow P_{\mathrm{goal}}^{t+1}$, $A_{\mathrm{goal}}^t \rightarrow A_{\mathrm{goal}}^{t+1}$ \\[1pt]
$A_{\mathrm{goal}}^{t+1} \rightarrow Count_{\mathrm{goal}}^{t+1}$}
&
\makecell[tl]{$P_{\mathrm{goal}}^{t+1} := P_{\mathrm{goal}}^t$ \\[1pt]
$A_{\mathrm{goal}}^{t+1} := A_{\mathrm{goal}}^t$ \\[1pt]
$Count_{\mathrm{goal}}^{t+1} := f_{\mathrm{count}}(A_{\mathrm{goal}}^{t+1})$}
\\[1pt]
\specialrule{0.03em}{2pt}{2pt}

\texttt{enemy > RandomNPC}
&
\makecell[tl]{$U_{\mathrm{enemy\_rng}}^t$, $P_{\mathrm{enemy}}^t$ \\[1pt]
$O_{\mathrm{enemy}}^t$, $V_{\mathrm{enemy}}^t$ \\[1pt]
$A_{\mathrm{enemy}}^t$, $P_{\mathrm{enemy}}^{t+1}$ \\[1pt]
$O_{\mathrm{enemy}}^{t+1}$, $V_{\mathrm{enemy}}^{t+1}$ \\[1pt]
$A_{\mathrm{enemy}}^{t+1}$}
&
\makecell[tl]{$U_{\mathrm{enemy\_rng}}^t \rightarrow P_{\mathrm{enemy}}^{t+1}$, $U_{\mathrm{enemy\_rng}}^t \rightarrow O_{\mathrm{enemy}}^{t+1}$ \\[1pt]
$U_{\mathrm{enemy\_rng}}^t \rightarrow V_{\mathrm{enemy}}^{t+1}$, $P_{\mathrm{enemy}}^t \rightarrow P_{\mathrm{enemy}}^{t+1}$ \\[1pt]
$O_{\mathrm{enemy}}^t \rightarrow P_{\mathrm{enemy}}^{t+1}$, $V_{\mathrm{enemy}}^t \rightarrow P_{\mathrm{enemy}}^{t+1}$ \\[1pt]
$A_{\mathrm{enemy}}^t \rightarrow A_{\mathrm{enemy}}^{t+1}$}
&
\makecell[tl]{$P_{\mathrm{enemy}}^{t+1} := f_{\mathrm{pos}}(P_{\mathrm{enemy}}^t, O_{\mathrm{enemy}}^t, V_{\mathrm{enemy}}^t, U_{\mathrm{enemy\_rng}}^t)$ \\[1pt]
$O_{\mathrm{enemy}}^{t+1} := f_{\mathrm{ori}}(O_{\mathrm{enemy}}^t, U_{\mathrm{enemy\_rng}}^t)$ \\[1pt]
$V_{\mathrm{enemy}}^{t+1} := f_{\mathrm{vel}}(V_{\mathrm{enemy}}^t, U_{\mathrm{enemy\_rng}}^t)$}
\\[1pt]
\midrule

\multicolumn{4}{l}{\textbf{LevelMapping + layout}} \\
\midrule
\texttt{LevelMapping + layout}
&
\makecell[tl]{$Init_c^0$, $P_s^t$ \\[1pt]
$A_s^t$, $c \mapsto s$}
&
\makecell[tl]{$Init_c^0 \rightarrow P_s^t$, $Init_c^0 \rightarrow A_s^t$}
&
\makecell[tl]{$P_s^t := \mathrm{init\_position}(Init_c^0)$ \\[1pt]
$A_s^t := \mathrm{init\_existence}(Init_c^0)$ \\[1pt]
for any layout symbol $c$ mapped to sprite $s$}
\\[1pt]
\midrule

\multicolumn{4}{l}{\textbf{InteractionSet}} \\
\midrule
\texttt{avatar wall > stepBack}
&
\makecell[tl]{$U_{\mathrm{action}}^t$, $P_{\mathrm{avatar}}^t$ \\[1pt]
$P_{\mathrm{wall}}^t$, $P_{\mathrm{avatar}}^{t+1}$}
&
\makecell[tl]{$U_{\mathrm{action}}^t \rightarrow P_{\mathrm{avatar}}^{t+1}$, $P_{\mathrm{avatar}}^t \rightarrow P_{\mathrm{avatar}}^{t+1}$ \\[1pt]
$P_{\mathrm{wall}}^t \rightarrow P_{\mathrm{avatar}}^{t+1}$}
&
\makecell[tl]{$P_{\mathrm{avatar}}^{t+1} := \mathrm{stepBack}(P_{\mathrm{avatar}}^t, P_{\mathrm{wall}}^t, U_{\mathrm{action}}^t)$}
\\[1pt]
\specialrule{0.03em}{2pt}{2pt}

\texttt{goal avatar > killSprite}
&
\makecell[tl]{$P_{\mathrm{goal}}^t$, $P_{\mathrm{avatar}}^t$ \\[1pt]
$A_{\mathrm{goal}}^t$, $A_{\mathrm{avatar}}^t$ \\[1pt]
$A_{\mathrm{goal}}^{t+1}$}
&
\makecell[tl]{$P_{\mathrm{goal}}^t \rightarrow A_{\mathrm{goal}}^{t+1}$, $P_{\mathrm{avatar}}^t \rightarrow A_{\mathrm{goal}}^{t+1}$ \\[1pt]
$A_{\mathrm{goal}}^t \rightarrow A_{\mathrm{goal}}^{t+1}$, $A_{\mathrm{avatar}}^t \rightarrow A_{\mathrm{goal}}^{t+1}$}
&
\makecell[tl]{$A_{\mathrm{goal}}^{t+1} := \mathrm{killSprite}(A_{\mathrm{goal}}^t,$ \\[1pt]
$\mathrm{overlap}(P_{\mathrm{goal}}^t, P_{\mathrm{avatar}}^t),$ \\[1pt]
$A_{\mathrm{avatar}}^t)$}
\\[1pt]
\specialrule{0.03em}{2pt}{2pt}

\texttt{avatar enemy > killSprite}
&
\makecell[tl]{$P_{\mathrm{avatar}}^t$, $P_{\mathrm{enemy}}^t$ \\[1pt]
$A_{\mathrm{avatar}}^t$, $A_{\mathrm{enemy}}^t$ \\[1pt]
$A_{\mathrm{avatar}}^{t+1}$}
&
\makecell[tl]{$P_{\mathrm{avatar}}^t \rightarrow A_{\mathrm{avatar}}^{t+1}$, $P_{\mathrm{enemy}}^t \rightarrow A_{\mathrm{avatar}}^{t+1}$ \\[1pt]
$A_{\mathrm{avatar}}^t \rightarrow A_{\mathrm{avatar}}^{t+1}$, $A_{\mathrm{enemy}}^t \rightarrow A_{\mathrm{avatar}}^{t+1}$}
&
\makecell[tl]{$A_{\mathrm{avatar}}^{t+1} := \mathrm{killSprite}(A_{\mathrm{avatar}}^t,$ \\[1pt]
$\mathrm{overlap}(P_{\mathrm{avatar}}^t, P_{\mathrm{enemy}}^t),$ \\[1pt]
$A_{\mathrm{enemy}}^t)$}
\\[1pt]
\midrule

\multicolumn{4}{l}{\textbf{TerminationSet}} \\
\midrule
\texttt{SpriteCounter stype=goal}
\newline
\texttt{limit=0 win=True}
&
\makecell[tl]{$A_{\mathrm{goal}}^{t+1}$, $Count_{\mathrm{goal}}^{t+1}$ \\[1pt]
$Terminal_{\mathrm{goal}}^{t+1}$}
&
\makecell[tl]{$A_{\mathrm{goal}}^{t+1} \rightarrow Count_{\mathrm{goal}}^{t+1}$,\\[1pt] $Count_{\mathrm{goal}}^{t+1} \rightarrow Terminal_{\mathrm{goal}}^{t+1}$}
&
\makecell[tl]{$Count_{\mathrm{goal}}^{t+1} := f_{\mathrm{count}}(A_{\mathrm{goal}}^{t+1})$ \\[1pt]
$Terminal_{\mathrm{goal}}^{t+1} := \mathbf{1}[Count_{\mathrm{goal}}^{t+1} \le 0]$}
\\[1pt]
\specialrule{0.03em}{2pt}{2pt}

\texttt{SpriteCounter stype=avatar}
\newline
\texttt{limit=0 win=False}
&
\makecell[tl]{$A_{\mathrm{avatar}}^{t+1}$, $Count_{\mathrm{avatar}}^{t+1}$ \\[1pt]
$Terminal_{\mathrm{avatar}}^{t+1}$}
&
\makecell[tl]{$A_{\mathrm{avatar}}^{t+1} \rightarrow Count_{\mathrm{avatar}}^{t+1}$,\\[1pt] $Count_{\mathrm{avatar}}^{t+1} \rightarrow Terminal_{\mathrm{avatar}}^{t+1}$}
&
\makecell[tl]{$Count_{\mathrm{avatar}}^{t+1} := f_{\mathrm{count}}(A_{\mathrm{avatar}}^{t+1})$ \\[1pt]
$Terminal_{\mathrm{avatar}}^{t+1} := \mathbf{1}[Count_{\mathrm{avatar}}^{t+1} \le 0]$}
\\[1pt]
\bottomrule
\end{tabularx}
\end{table*}

The transition from $t$ to $t+1$ is defined by structural functions:
\begin{equation}
    X_i^{t+1} := f_i(\mathrm{Pa}(X_i^{t+1})),
\end{equation}
where $\mathrm{Pa}(X_i^{t+1})$ denotes the causal parents of $X_i^{t+1}$. 
These parents are not learned from correlations. 
They are obtained from the VGDL elements: \texttt{SpriteSet}, \texttt{LevelMapping}, \texttt{InteractionSet}, and \texttt{TerminationSet}. 
\Cref{tab:basicgame-scm} gives the mapping for \texttt{basicgame}.

\subsection{Sprite Dynamics}
\label{subsec:sprite-dynamics}

Each sprite class contributes a local causal mechanism. 
These mechanisms are derived from the sprite definitions in the \texttt{SpriteSet} of the VGDL description shown in \Cref{fig:basicgame-complete}. 
In the mapping in \Cref{tab:basicgame-scm}, sprite dynamics define how attributes such as position, orientation, velocity, and alive-state are updated from time $t$ to time $t+1$.

For a moving sprite $s$, the next position is caused by its current position, orientation, velocity, and the relevant input:
\begin{equation}
    P_s^{t+1} :=
    f_{\mathrm{pos}}(P_s^t, O_s^t, Vel_s^t, U_s^t).
\end{equation}
For the avatar, $U_s^t = U_{\mathrm{action}}^t$. 
For a random enemy, $U_s^t = U_{\mathrm{enemy\_rng}}^t$. 
The orientation and velocity updates are
\begin{equation}
    O_s^{t+1} :=
    f_{\mathrm{ori}}(O_s^t, U_s^t), \quad Vel_s^{t+1} :=
    f_{\mathrm{vel}}(Vel_s^t, U_s^t).
\end{equation}

For immovable sprites such as walls and goals, the position mechanism is an identity function: $P_s^{t+1} := P_s^t$.
Their alive-state persists unless an interaction rule changes it: $    A_s^{t+1} := A_s^t$.

This distinction makes the causal role of sprite classes explicit. 
A \texttt{MovingAvatar} has an action-dependent movement mechanism, a \texttt{RandomNPC} has an RNG-dependent movement mechanism, and an \texttt{Immovable} sprite has a persistence mechanism. 
The sprite dynamics, therefore, define the baseline state transition before interaction rules such as \texttt{stepBack} or \texttt{killSprite} modify the next state.

\subsection{Interaction Mechanisms}
\label{subsec:interaction-mechanisms}

Interaction mechanisms are derived from the \texttt{InteractionSet} in the VGDL description. 
They define how the state of one sprite changes when it spatially interacts with another sprite. 
In the mapping in \Cref{tab:basicgame-scm}, these mechanisms include rules such as \texttt{stepBack} and \texttt{killSprite}.
For two sprites $i$ and $j$, an interaction is activated by a spatial condition:
\begin{equation}
    C_{ij}^{t} := \mathrm{overlap}(P_i^t, P_j^t).
\end{equation}
The resulting state update can be written as
\begin{equation}
    X_i^{t+1} :=
    f_{\mathrm{int}}(X_i^t, P_i^t, P_j^t, A_i^t, A_j^t, C_{ij}^{t}),
\end{equation}
where $X_i^{t+1}$ is the affected next-state variable.

For example, the rule \texttt{avatar wall > stepBack} changes the avatar's next position when the attempted movement causes contact with a wall:
\begin{equation}
    P_{\mathrm{avatar}}^{t+1} :=
    \mathrm{stepBack}
    (P_{\mathrm{avatar}}^t, P_{\mathrm{wall}}^t, U_{\mathrm{action}}^t).
\end{equation}
Similarly, \texttt{killSprite} changes the alive-state of the affected sprite when the required overlap condition holds:
\begin{equation}
    A_i^{t+1} :=
    \mathrm{killSprite}(A_i^t, C_{ij}^{t}, A_j^t).
\end{equation}

Thus, interaction mechanisms specify causal effects of spatial contact. 
They do not merely record that two sprites are close to each other. 
They define which sprite state is changed, under which spatial condition, and by which VGDL rule.

\subsection{Termination Mechanisms}
\label{subsec:termination-mechanisms}

Termination mechanisms are derived from the \texttt{TerminationSet} in the VGDL description. 
They define when the game reaches a win or loss state. 
In the mapping in \Cref{tab:basicgame-scm} for \texttt{basicgame}, termination depends on live-instance counts for sprites such as the goal and the avatar.

For a sprite type $s$, the live-instance count is computed from the alive-state variables:
\begin{equation}
    Count_s^{t+1} :=
    f_{\mathrm{count}}(A_s^{t+1}).
\end{equation}
A terminal variable is then computed from this count:
\begin{equation}
    Terminal_s^{t+1} :=
    \mathbf{1}[Count_s^{t+1} \leq limit_s].
\end{equation}

For example, in \texttt{basicgame}, the game is
\begin{itemize}
\item won when no goal sprites remain: \\ $\hphantom{.}\qquad \quad~
    Terminal_{\mathrm{goal}}^{t+1} :=
    \mathbf{1}[Count_{\mathrm{goal}}^{t+1} \leq 0].
$
\item lost when no avatar sprites remain: \\
$\hphantom{.}\qquad \quad Terminal_{\mathrm{avatar}}^{t+1} :=
    \mathbf{1}[Count_{\mathrm{avatar}}^{t+1} \leq 0].
$
\end{itemize}

Termination nodes are, therefore, leaf nodes in the SCM. 
They summarize the consequences of earlier sprite dynamics and interaction mechanisms, but they do not directly cause movement or collision outcomes within the same transition.





\section{Discussion and Future Work}

This dynamic SCM framework has several benefits. It provides \emph{explainability} by making causal pathways explicit, allows \emph{simulation of counterfactuals}, and supports \emph{causal RL} by supplying a full state model to agents. It also enables \emph{procedural content validation} by testing whether a newly generated game description is logically consistent through its implied causal model, as done by Jiwatode et al.~\cite{jiwatode2026gameplay} for VGDL synthesis. 

One of the primary limitations of this approach is that the causal model's size scales with the complexity of the VGDL description. Additionally, games with intricate physics or hidden states may require structural approximations. Furthermore, if the actual game implementation deviates from the VGDL design (e.g., due to bugs or unmodeled nuances), errors will naturally arise in the causal mapping.

Future work should address these issues by developing semi-automated tools to extract SCMs directly from game code and by extending the framework to multi-agent or real-time games. This framework could also be extended to other games in the GVGAI suite and to other game definitions. Future research should also empirically evaluate these SCMs to quantify their impact on agent performance. Different methods to encode spatial relationships in the causal model could also be explored.

In conclusion, by mapping video games to a dynamic causal framework, we enable reasoning about actions and outcomes in games, complementing statistical and learning-based methods.

\subsection*{Acknowledgements:}  This work is jointly supported by the “HybrInt - Hybrid Intelligence through Interpretable AI in Machine Perception and Interaction” project (Zukunft Nds, Niedersächsisches Ministerium für Wissenschaft, Grant ID: ZN4219) and the Novo Nordisk Foundation grant NNF25OC0105856. 

The editing process was supported by Google’s Gemini.

\bibliographystyle{IEEEtran}
\bibliography{references}

@inproceedings{jiwatode_spatial_reasoning,
  title={Spatial Reasoning in LLM Game Agents: Impact of Causal Context and Multi-Step Planning},
  author={Jiwatode, Mohit and Fuchs, Ronja and Schmöcker, Robin and Rosenhahn, Bodo and Dockhorn, Alexander},
  booktitle={2026 IEEE Conference on Games (CoG)},
  pages={1--8},
  year={2026},
  organization={IEEE}
}

@article{jiwatode2026gameplay,
  title={From Gameplay Traces to Game Mechanics: Causal Induction with Large Language Models},
  author={Jiwatode, Mohit and Dockhorn, Alexander and Rosenhahn, Bodo},
  journal={arXiv:2602.00190},
  year={2026}
}

@article{li2025gvgai,
  title={GVGAI-LLM: Evaluating Large Language Model Agents with Infinite Games},
  author={Li, Yuchen and Lin, Cong and Nasir, Muhammad Umair and Bontrager, Philip and Liu, Jialin and Togelius, Julian},
  journal={arXiv:2508.08501},
  year={2025}
}

@inproceedings{madumal2020explainable,
  title={Explainable reinforcement learning through a causal lens},
  author={Madumal, Prashan and Miller, Tim and Sonenberg, Liz and Vetere, Frank},
  booktitle={Proceedings of the AAAI conference on artificial intelligence},
  volume={34},
  number={03},
  pages={2493--2500},
  year={2020}
}

@book{gvgaibook2019,
    title={General Video Game Artificial Intelligence},
    author={Diego Perez-Liebana and Simon M. Lucas and Raluca D. Gaina and Julian Togelius and Ahmed Khalifa and Jialin Liu},
    journal={Synthesis Lectures on Games and Computational Intelligence},
    volume={3},
    number={2},
    pages={1--191},
    publisher={Morgan \& Claypool Publishers},
    note={\url{https://gaigresearch.github.io/gvgaibook/}},
    year={2019}
}

@article{chen2025causal,
  title={Causal-aware Large Language Models: Enhancing Decision-Making Through Learning, Adapting and Acting},
  author={Chen, Wei and Zhang, Jiahao and Zhu, Haipeng and Xu, Boyan and Hao, Zhifeng and Zhang, Keli and Ye, Junjian and Cai, Ruichu},
  journal={arXiv:2505.24710},
  year={2025}
}

@article{hammond2023reasoning,
  title={Reasoning about causality in games},
  author={Hammond, Lewis and Fox, James and Everitt, Tom and Carey, Ryan and Abate, Alessandro and Wooldridge, Michael},
  journal={Artificial Intelligence},
  volume={320},
  pages={103919},
  year={2023},
  publisher={Elsevier}
}

@book{pearl2009causality,
  title={Causality},
  author={Pearl, Judea},
  year={2009},
  publisher={Cambridge University Press}
}

@article{deng2023causal,
  title={Causal reinforcement learning: A survey},
  author={Deng, Zhihong and Jiang, Jing and Long, Guodong and Zhang, Chengqi},
  journal={arXiv preprint arXiv:2307.01452},
  year={2023}
}

@article{li2026confounding,
  title={Confounding robust deep reinforcement learning: A causal approach},
  author={Li, Mingxuan and Zhang, Junzhe and Bareinboim, Elias},
  journal={Advances in Neural Information Processing Systems},
  volume={38},
  pages={138292--138325},
  year={2026}
}

@article{Apeldoorn2021Exception,
  author = {Apeldoorn, Daan and Dockhorn, Alexander},
  title = {Exception-Tolerant Hierarchical Knowledge Bases for Forward Model Learning},
  journal = {IEEE Transactions on Games},
  year = {2021},
  pages = {1–14},
}

@inproceedings{smith2010ludocore,
  title={Ludocore: A logical game engine for modeling videogames},
  author={Smith, Adam M and Nelson, Mark J and Mateas, Michael},
  booktitle={Proceedings of the 2010 IEEE Conference on Computational Intelligence and Games},
  pages={91--98},
  year={2010},
  organization={IEEE}
}

@incollection{genesereth2022analyzing,
  title={Analyzing Games with Logic},
  author={Genesereth, Michael and Thielscher, Michael},
  booktitle={General Game Playing},
  pages={129--149},
  year={2022},
  publisher={Springer}
}

\clearpage

\end{document}